\documentclass{article}
\usepackage{amsmath,epsfig}
\usepackage[preprint]{spconfa4}
\usepackage{xcolor}
\usepackage{cite}

\usepackage{multirow}
\usepackage{pifont}
\usepackage{amsfonts,amssymb}

\let\OLDthebibliography\thebibliography
\renewcommand\thebibliography[1]{
  \OLDthebibliography{#1}
  \setlength{\parskip}{0pt}
  \setlength{\itemsep}{0pt plus 0.3ex}
}

\begin{document}\sloppy

\def\x{{\mathbf x}}
\def\L{{\cal L}}

\title{Locality-aware Attention Network with Discriminative Dynamics Learning for weakly supervised anomaly detection}
%
\name{Yujiang Pu$^{1}$, Xiaoyu Wu$^{1\ast}$ }
\address{$^{1}$State Key Laboratory of Media Convergence and Communication,\\Communication University of China, Beijing, China (*Corresponding author) \\ \{pyj2020, wuxiaoyu\}@cuc.edu.cn}

\maketitle

\begin{abstract}
Video anomaly detection is recently formulated as a multiple instance learning task under weak supervision, in which each video is treated as a bag of snippets to be determined whether contains anomalies. Previous efforts mainly focus on the discrimination of the snippet itself without modeling the temporal dynamics, which refers to the variation of adjacent snippets. Therefore, we propose a Discriminative Dynamics Learning (DDL) method with two objective functions, i.e., dynamics ranking loss and dynamics alignment loss. The former aims to enlarge the score dynamics gap between positive and negative bags while the latter performs temporal alignment of the feature dynamics and score dynamics within the bag. Moreover, a Locality-aware Attention Network (LA-Net) is constructed to capture global correlations and re-calibrate the location preference across snippets, followed by a multilayer perceptron with causal convolution to obtain anomaly scores. Experimental results show that our method achieves significant improvements on two challenging benchmarks, i.e., UCF-Crime and XD-Violence.

\end{abstract}
\begin{keywords}
Video anomaly detection, weak supervision, multiple instance learning, temporal dynamics learning
\end{keywords}
\section{Introduction}
\label{sec:intro}

In recent years, the popularity of surveillance equipment has played an important role in preventing criminal behavior and maintaining public security. Relying on manual efforts to monitor these videos can no longer meet the urgent practical needs. Therefore, autonomous anomaly detection has received high attention due to its wide range of promising applications, including violence detection, traffic monitoring, content assessment, etc.

Given that anomalies are relatively scarce in real-life occasions, most previous work focuses on unsupervised algorithms \cite{chan2008ucsd,lu2013abnormal,ravanbakhsh2017abnormal,bergmann2020uninformed}, which learn the frequently occurring events as normal using one-class classifiers. Anomalies are then identified as outliers according to their departure from the learned representations of the normal class \cite{sabokrou2018adversarially,nguyen2019anomaly,gong2019memorizing,zaheer2020old}. However, it is infeasible to collect all kinds of normal scenarios in real cases, and there is a huge intraclass diversity among normal classes, which may lead to incorrect detection of emerging normal categories.

\begin{figure}[t]
	\centering
	\includegraphics[width=\linewidth]{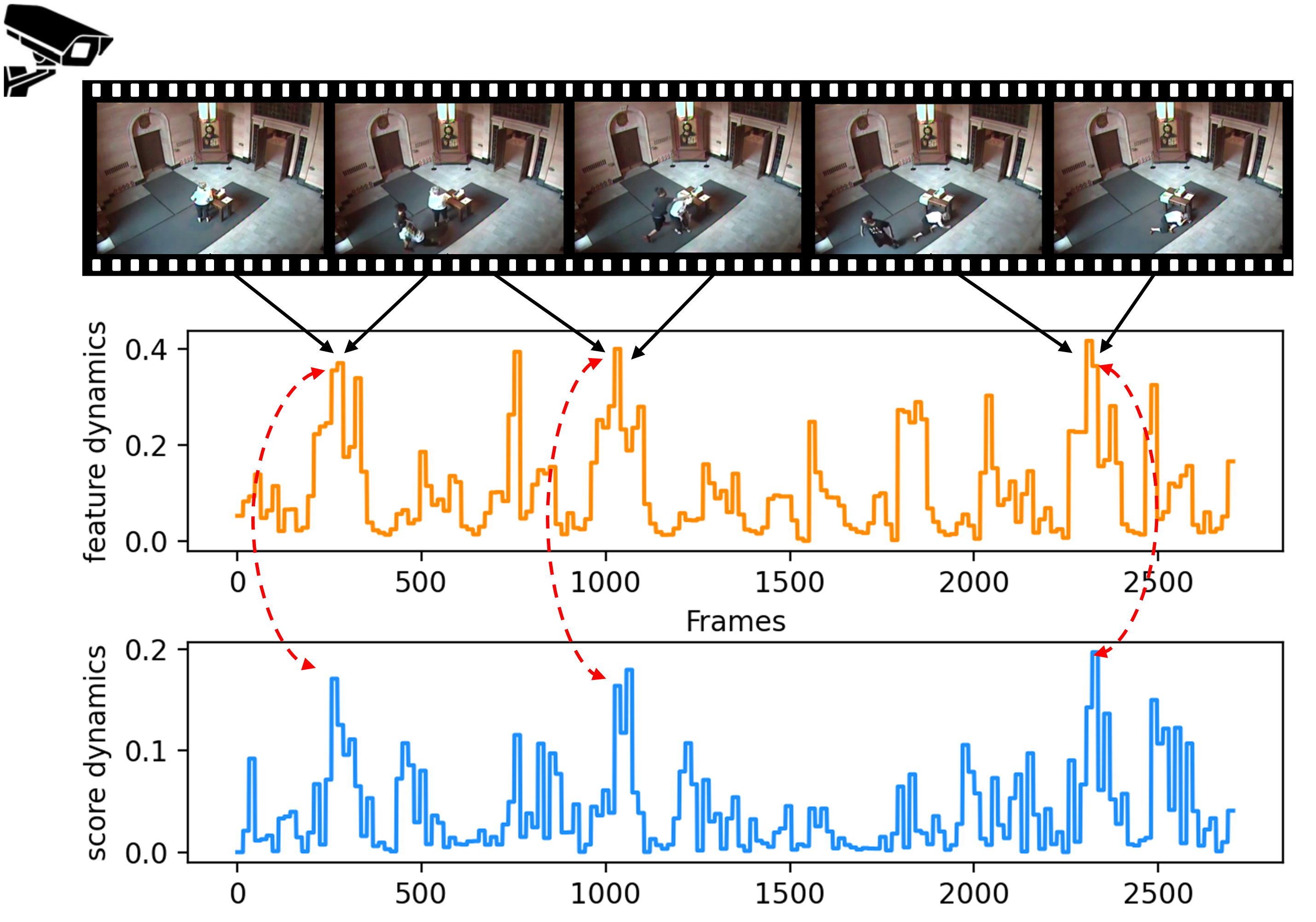}
	\caption{Temporal consistency of feature dynamics and score dynamics. The peak of score dynamics caused by feature dynamics generally indicates a sharp difference in the current frame, i.e., a sudden intrusion, which further suggests a potential abnormal event.}
\end{figure}

Another widely used paradigm for anomaly detection is weakly supervised learning~\cite{sultani2018real,zhang2019temporal,zhong2019graph}. In this case, all events in a normal video are marked as normal while those in abnormal videos are treated as anomalous, even if some events are actually normal. Such methods reduce the cost of annotations, but these noisy labels present a new challenge for anomaly detection. Sultani et al.~\cite{sultani2018real} propose a multiple instance learning approach to detect anomalies, where each video is regarded as a bag consisting of consecutive snippets. Anomaly videos are treated as positive bags, and normal ones are seen as negative bags. A multiple ranking loss with two constraints is introduced to distinguish positive and negative bags. In~\cite{wan2020weakly}, Wan et al. construct a center regression loss to smooth anomaly scores while aggregating the activation of normal samples. Wu et al.~\cite{wu2020not} develop an HL-Net to capture both long-range dependencies and local distance relations with auxiliary audio information. A MIST framework is built in~\cite{feng2021mist} to generate pseudo labels for anomalous samples and employ a self-guided attention module to encode anomalous regions.

However, the above methods specialize in the discrimination of the snippets themselves without delving into the variation of adjacent snippets, which we call temporal dynamics. We observe that most anomalies typically occur at locations where the temporal dynamics changes dramatically, and such changes include both feature dynamics and score dynamics. As shown in Fig.\,1, if abnormal behavior suddenly appears in the current frame, its corresponding feature dynamics usually change sharply, further leading to a large difference in the anomaly scores. The score dynamics in the positive bag should be greater than that in the negative bag, and the two types of temporal dynamics are shown to be causally consistent over the temporal dimension.

To address the above issues, we propose a novel Discriminative Dynamics Learning (DDL) method including two loss functions, of which the dynamics ranking loss captures the variation magnitude of anomalies by enlarging the accumulated score dynamics between positive and negative bags while the dynamics alignment loss achieves implicit ordering by aligning feature dynamics with score dynamics over the temporal dimension. Besides, a Locality-aware Attention Network (LA-Net) is built to model the long-range dependencies and re-calibrate the locality preference of adjacent snippets. Extensive  experiments on two challenging benchmark demonstrate the superiority of our proposed framework over the current state-of-the-art methods.

\section{proposed method}

\begin{figure*}[h]
	\centering
	\includegraphics[width=\linewidth]{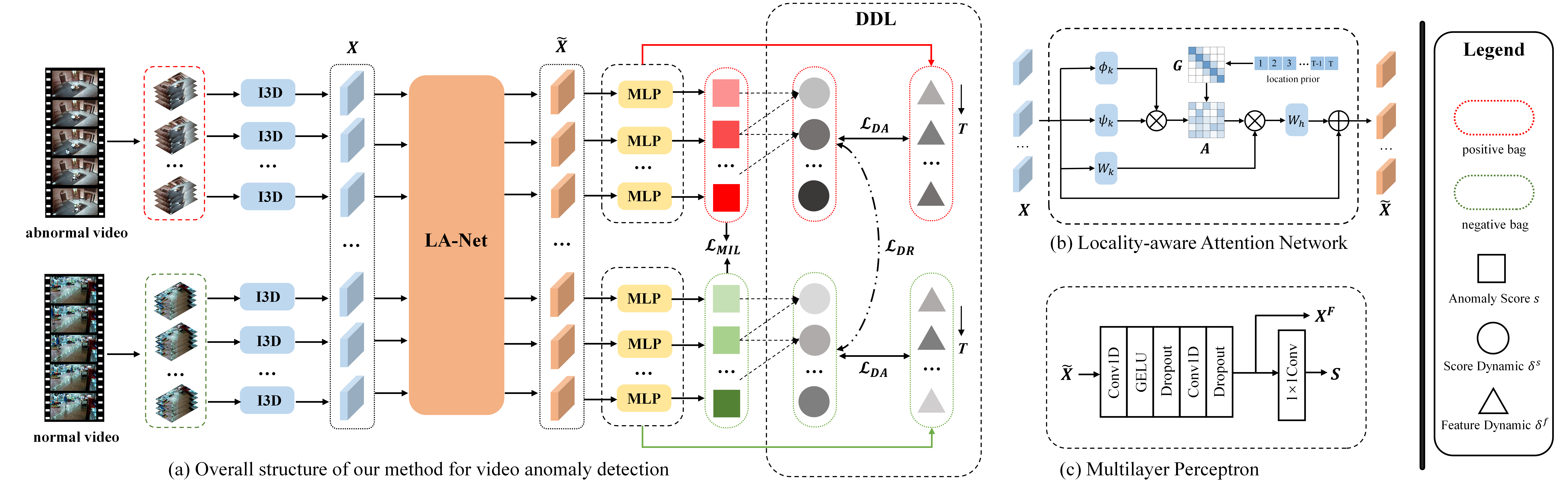}
	\caption{Overview of the proposed video anomaly detection framework. It comprises a Locality-aware Attention Network (LA-Net), a Multilayer Perceptron (MLP) and Discriminative  Dynamics Learning (DDL).}
\end{figure*}

\subsection{Formulation}
In this paper, we consider video anomaly detection as a multiple instance learning task under weak supervision. Specifically, a video snippet bag $\mathcal{X}=\{x_i\}^T_{i=1}$ and its corresponding video-level label $y\in\{0,1\}$ are given, where $y=0$ indicates that the current video is normal otherwise  $y=1$ means that there exists anomalies in the video. In the test stage, the predicted anomaly score of the video is expressed as $\mathcal{S}=\{s_j\}^T_{j=1}$, where $s_j$ is the anomaly score of the $j^{th}$ snippet. The goal of anomaly detection is to determine whether the video contains anomalies based on the predicted anomaly score while pinpointing the interval.

The overall structure of our method is shown in Fig.\,2(a). Formally, a long untrimmed video is first divided into non-overlapping snippets through a sliding window of 16 frames. Then, we use a pre-trained I3D network~\cite{carreira2017quo} to extract snippet features and concatenate them over the temporal dimension to obtain video feature $X\in\mathbb{R}^{T\times D}$, where $T$ is the number of snippets and $D$ is the feature dimension. Subsequently, the LA-Net is used to model the global dependencies and re-calibrate the local preference. The calibrated feature $\widetilde{X}$ is then fed into a Multilayer Perceptron (MLP) to capture robust snippet representation $X^F$, followed by a causal convolution layer to obtain the anomaly score $\mathcal{S}$. A Multiple Instance Learning (MIL) loss is used to monitor the discrimination of the snippets. The robust representation and the anomaly scores are further utilized in DDL for auxiliary optimization.

\subsection{Locality-aware Attention Network}

The Locality-aware Attention Network (LA-Net) aims to model the long-range  dependencies across snippets while re-calibrate the contextual relations of local distance, as shown in Fig.\,2(b). Following the self-attention mechanism~\cite{vaswani2017attention}, we construct the global attention map as

\begin{equation}
	\mathcal{A}_{ij} = \frac{exp\{\mathcal{R}_k(x_i,x_j)\}}{\sum_{n=1}^Texp\{\mathcal{R}_k(x_i,x_n)\}},
\end{equation}

\noindent where $\mathcal{R}_k=\phi(x_i)^T\psi(x_j)$ is a multi-head kernel function to capture diversity semantic patterns, $\phi(\cdot)$ and $\psi(\cdot)$ are two linear functions applied on input snippets, and $k$ is the head numbers. 

The generated global attention map develops the long-distance relation across snippets but ignores the inherent location preference of the video sequence. Thus, a Gaussian-like location prior is introduced to re-calibrate the local contextual correlations, which can be expressed as

\begin{equation}
	\mathcal{G}_{ij} = exp(-\frac{|i-j|^2}{2\sigma}),
\end{equation}

\begin{equation}
	\widetilde{\mathcal{A}}=\mathcal{A}+\mathcal{G},
\end{equation}

\noindent where $i$ and $j$ are the relative positions of the snippets, and $\sigma$ is a hyperparamter controlling the deviation of the center location. Generally, a higher $\mathcal{G}_{ij}$ indicates a higher prior probability of the $i^{th}$ snippet to the $j^{th}$ snippet. By adding $\mathcal{G}$ as a bias term to the global attention map, the representation of the current snippet can be explicitly enhanced while suppressing the interference of long-distance. Correspondingly, our Locality-aware Attention Network is formulated as

\begin{equation}
	\widetilde{X} = Norm(stack(\widetilde{\mathcal{A}}XW_k)W_h+X),
\end{equation}

\noindent where $W_k\in\mathbb{R}^{D\times\frac{D_h}{k}}$ and $W_h \in \mathbb{R}^{D_h \times D}$ are two linear projection layers, and $Norm(\cdot)$ denotes the layer normalization.

\subsection{Multiple Instance Learning}
Subsequently, a two-layered MLP $\mathcal{F}^\Theta$ with GELU activation and dropout operation is followed to obtain robust snippet representation $X^F$, as shown in Fig.\,2(c). To acquire reliable anomaly scores, we use a $1\times1$ causal convolution layer to capture historical observations. The process is expressed as

\begin{equation}
	\mathcal{S} = \sigma(W\mathcal{F}^\Theta(\widetilde{X})+b),
\end{equation}

\noindent where $W$ is a $1\times1$ convolution with kernel size $K$ and $b$ is a bias term. $\sigma(\cdot)$ denotes a sigmoid function and $\mathcal{S}\in\mathbb{R}^{1\times T}$ is the temporal anomaly scores of the video bag.

Following~\cite{wan2020weakly,wu2020not,paul2018w}, we use the $k$-max approach to calculate multiple instance learning (MIL) loss. Specifically, if the video is abnormal, the average of the top-$k$ prediction results in $S$ is used as the video anomaly score, where $ k=\lfloor\frac{T}{16}+1\rfloor$. If the video is normal then the highest prediction score is selected, i.e., $ k=1$. Accordingly, our MIL-based loss function is formulated as

\begin{equation}
	\mathcal{L}_{MIL} = \frac{1}{N} \sum^N_{i=1} -y_i log({p}_i),
\end{equation}

\noindent where $\{{p}_i\}^N_{i=1}$ is the video anomaly score and $\{{y}_i\}^N_{i=1}$ is the binary video-level ground truth.

\subsection{Discriminative Dynamics Learning}

Since the anomaly score in the negative bag is usually close to the positive one at the early stage of training, direct score ranking patterns~\cite{sultani2018real,zhang2019temporal,zhen2021multi} are likely to produce incorrect decision boundaries, which may lead to further deviations from the ground truth. To alleviate this problem, we design a dynamics ranking loss that amplifies the magnitude of abnormal score changes while suppressing score fluctuations.

Specifically, the anomaly score of a video bag is denoted as $ S=\{s_1,s_2,\ldots,s_t\} $ and its corresponding score dynamics is expressed as $ \Delta S = \{\delta^s_1,\delta^s_2,\ldots,\delta^s_{t-1}\} $, where $ \delta^s_t=|s_t-s_{t+1}|$. Then, we use the top-$k$ dynamics in the bag to calculate the accumulation of score dynamics, which is expressed as

\begin{equation}
	\mathcal{E}_{\Delta S} = \frac{1}{k}\sum^k_{t=1}\mid \delta^s_t\mid^2,
\end{equation}

\noindent where $k=\lfloor\frac{T}{16}+1\rfloor$ if the video is abnormal otherwise $k=1$. Considering that the dynamics accumulation of a positive bag should be remarkably larger than that of a negative bag, the dynamics ranking (DR) loss is naturally formulated as

\begin{equation}
	\mathcal{L}_{DR} = max(0,\zeta-\mathcal{E}^a_{\Delta S}+\mathcal{E}^n_{\Delta S}),
\end{equation}

\noindent where $\mathcal{E}^a_{\Delta S}$ and $\mathcal{E}^n_{\Delta S}$ are dynamics accumulation of positive and negative bags, respectively, and $\zeta$ is a margin hyperparameter. This loss function enlarges the dynamics gap between positive and negative bags, further leading to a suppression of dynamics accumulation in the negative bag while increasing that of the positive bag.

Besides, snippets at abnormal boundaries are more variable, resulting in a sharp jittering at the junction. This jittering is reflected in the change of anomaly scores after forward discrimination, that is, the score dynamics and feature dynamics have a causal consistency over the temporal dimension. To this end, we introduce a dynamics alignment strategy to model this tendency. Given the robust snippet representation after two-layered MLP denoted as $ X^F=\{x^F_1,x^F_2,\ldots,x^F_t\}$, its corresponding feature dynamics can be expressed as $ \Delta X = \{\delta^f_1,\delta^f_2,\ldots,\delta^f_{t-1}\} $, in which the $\delta^f_t$ is calculated as

\begin{equation}
	\delta^f_t=1-\frac{x^F_tx^F_{t+1}}{\left \|x^F_t\right\|\left \|x^F_{t+1}\right\|},
\end{equation}

Considering that the feature dynamics and the score dynamics are in two different semantic spaces, we use Kullback-Leibler (KL) divergence rather than mean square error to learn this property. Thus, our dynamics alignment (DA) loss is formulated as follows:

\begin{equation}
	\mathcal{L}_{DA} = \frac{1}{N\times(T-1)}\sum^N_{i=1}(\sum^{T-1}_{t=1}-\delta^s_tlog(\delta^f_t+\epsilon))_i,
\end{equation}

\noindent where $\{\delta^s_t\}^{T-1}_{t=1}$ and $\{\delta^f_t\}^{T-1}_{t=1}$ are score dynamics and feature dynamics in the same video bag, respectively, and $\epsilon$ is a very small scalar to prevent feature dynamics from being zero. This loss makes the feature dynamics distribution consistent with the score dynamics over temporal dimension, while implicitly increasing the discriminability of normal and abnormal snippets within the video bag.

Combined with the MIL-based loss, the overall objective function of our  model is formulated as

\begin{equation}
\begin{split}
    \mathcal{L} = \mathcal{L}_{MIL}+\lambda_1\mathcal{L}_{DR}+\lambda_2\mathcal{L}_{DA},
\end{split}
\end{equation}

\noindent where $\lambda_1$ and $\lambda_2$ are pre-defined weights for dynamics ranking loss and dynamics alignment loss, respectively.

\section{experiments}

In this section, we evaluate our approach on two benchmark datasets (UCF-Crime and XD-Violence). First, we present the two datasets and their corresponding evaluation metrics in detail. The implementation details of the experiments are then given. Finally, we compare our results with several state-of-the-art methods, a series of quantitative and qualitative experimental results elucidating the effectiveness of our method. 

\subsection{Datasets}

\textbf{UCF-Crime} \cite{sultani2018real} is a large-scale dataset collected from surveillance with a total duration of 128 hours, containing 1900 long untrimmed videos. The dataset covers 13 types of anomalies in 1610 training videos and 290 test videos, where the training videos have only video-level labels while the test provides frame-level annotations. Following~\cite{sultani2018real,liu2019exploring,wan2020weakly,feng2021mist}, we use the area under the frame-level ROC curve (AUC) to evaluate the performance of the proposed method.

\noindent\textbf{XD-Violence} \cite{wu2020not} is a newly released anomaly detection dataset that covers 6 types of violence events. The dataset consists of 4754 untrimmed videos including video-level labeled training set and frame-level labeled test set, with a total duration of 217 hours. Unlike UCF-Crime, the dataset is collected from various scenes including movies, sports, surveillance footage, CCTV, etc. As in \cite{wu2020not,tian2021weakly}, we use frame-level average precision (AP) to evaluate the proposed method.

\subsection{Implementation Details}

The head number $k$ of LA-Net is set to 4, and the hidden dimension $D_h$ for UCF-Crime is set to 512, which is adopted as 128 for XD-Violence. The location prior value $\sigma$ is empirically set to 16 and 6 for UCF-Crime and XD-Violence, respectively. The two Conv1D layers in MLP have 512 nodes and 128 nodes, regularized by dropout with a probability of 0.1 between each layer. The kernel sizes of the $1\times1$ causal convolution are set to 10 and 5 for UCF-Crime and XD-violence, respectively. Limited by computational resources, we use a uniform sampling strategy as in~\cite{wu2020not,zhen2021multi} to obtain a fixed length of 200 snippets in the training phase.

For a balanced purpose, the weights $\lambda_1$ for UCF-Crime and XD-violence are set to 1 and 2, respectively, while $\lambda_2$ is adopted as 1. The positive scalar $\epsilon$ is set to $10^{-7}$, and the margin $\zeta$ is set to 0. We use the Adam optimizer with a 128 mini-batch size to train the model for 50 epochs, and the initial learning rate is set to $5\times10^{-4}$ with a cosine decay strategy. All experiments are conducted on an NVIDIA Tesla A40 GPU based on PyTorch.

\subsection{Quantitative Comparison with existing Methods}

We first compare our approach against several state-of-the-art methods under weak supervision. Table 1 presents the frame-level AUC values on the UCF-Crime dataset. Notably, our method surpasses the previous state-of-the-art MIL-based methods,  Sultani et al.~\cite{sultani2018real} by 7.2\%, Zhong et al.~\cite{zhong2019graph} by 3\%, Wu et al.~\cite{wu2020not} by 2.68\%. Even when compared to the latest RTFM~\cite{tian2021weakly}, our result is also ahead by 0.82\%. 

Table 2 shows the frame-level AP values on the XD-Violence dataset. Remarkably, our approach substantially outperforms other methods, including both unsupervised and weakly supervised methods. Our result is 5.31\% higher than Wu et al.~\cite{wu2020not} and exceeds RTFM~\cite{tian2021weakly} by 2.91\%, reaching a new state-of-the-art frame-level AP of 80.72\%. Moreover, our method is the only one exceeding 80\% in terms of AP on XD-Violence with only RGB features. The superior results on both datasets demonstrate the effectiveness and capacity of the proposed method.

Compared with previous approaches, our LA-Net introduces explicit location prior to model local contextual relations, which suppresses long-distance redundancies and strengthens the local representation of adjacent snippets. Moreover, the proposed DDL method effectively captures the temporal dynamics of the video bag. The accumulation of score dynamics is ranked to increase the discriminability between positive and negative bags, while the feature dynamics are aligned with score dynamics within the bag to achieve causal consistency over the temporal dimension.

\begin{table}[h]
\centering
\caption{Frame-level AUC performance on UCF-Crime.}
\begin{tabular}{c|c|c} 
\hline
Method                 & Feature  & AUC(\%)         \\ 
\hline
Sultani \textit{et al.}~\cite{sultani2018real}                & C3D RGB  & 75.41           \\
Zhang \textit{et al.}~\cite{zhang2019temporal}                  & C3D RGB  & 78.66           \\
Motion-Aware~\cite{zhu2019motion}           & PWC Flow & 79.00           \\
Zhong \textit{et al.}~\cite{zhong2019graph}                  & TSN RGB  & 82.12           \\
Wu \textit{et al.}~\cite{wu2020not}                     & I3D RGB  & 82.44           \\
MS-BSAD~\cite{zhen2021multi}                & I3D RGB  & 83.53           \\
RTFM~\cite{tian2021weakly}                   & I3D RGB  & 84.30           \\
\textbf{DDL (Ours)} & I3D RGB  & \textbf{85.12}  \\
\hline
\end{tabular}
\end{table}

\begin{table}[h]
\centering
\caption{Frame-level AP performance on XD-Violence.}
\begin{tabular}{c|c|c} 
\hline
Method                 & Feature & AP(\%)          \\ 
\hline
SVM baseline           & -       & 50.78           \\
OCSVM~\cite{scholkopf1999support}                  & -       & 27.25           \\
Hasan \textit{et al.}~\cite{hasan2016learning}                  & -       & 30.77           \\ 
\hline
Sultani \textit{et al.}~\cite{sultani2018real}                & C3D RGB & 73.20           \\
Wu \textit{et al.}~\cite{wu2020not}                     & I3D RGB & 75.41           \\
RTFM~\cite{tian2021weakly}                   & I3D RGB & 77.81           \\
\textbf{DDL (Ours)} & I3D RGB & \textbf{80.72}  \\
\hline
\end{tabular}
\end{table}

\subsection{Ablation Study}

 The location prior is a critical component of our LA-Net. Table 3 shows that the introduction of the prior improves the performance of the model on both datasets, with a 0.61\% improvement on UCF-Crime and a 0.77\% increase on XD-Violence. This indicates that video sequences have a certain location preference, which can be enhanced by adding a specific location prior. Meanwhile, the location prior can further suppresses the long-distance noise information and improve the robustness of the current snippet, which has a positive impact on modeling adjacent temporal dynamics.

We then compare the effect of different dynamics loss functions, as shown in Table 4. Both $\mathcal{L}_{DR}$ and $\mathcal{L}_{DA}$ loss have distinct improvements on the basis of $\mathcal{L}_{MIL}$. Notably, the effect of $\mathcal{L}_{DA}$ is better than that of $\mathcal{L}_{DR}$, which illustrates the importance of maintaining causal consistency of temporal dynamics within the bag. Moreover, the synergistic effect of both is demonstrated when introduced simultaneously, which are shown to be complementary with the MIL-based loss. In addition, the feature dynamics produce an implicit ordering in the process of alignment with the score dynamics, which further enhances the discriminability of the video snippets.

\begin{table}[h]
\centering
\caption{Ablation study of location prior.}
\begin{tabular}{ccc} 
\hline
\multirow{2}{*}{Model}                       & UCF-Crime & XD-Violence  \\
                                              & AUC(\%)   & AP(\%)       \\ 
\hline
\multicolumn{1}{l}{LA-Net w/o prior $\mathcal{G}$}  & 83.06     & 78.41        \\
\multicolumn{1}{l}{LA-Net w/ prior $\mathcal{G}$} & 83.67     & 79.18        \\
\hline
\end{tabular}
\end{table}

\begin{table}[h]
\centering
\caption{Ablation study of the DDL method.}
\begin{tabular}{ccccc} 
\hline
\multirow{2}{*}{$\mathcal{L}_{MIL}$} & \multirow{2}{*}{$\mathcal{L}_{DR}$} & \multirow{2}{*}{$\mathcal{L}_{DA}$} & UCF-Crime & XD-Violence  \\
                      &                       &                       & AUC(\%)   & AP(\%)       \\ 
\hline
$\checkmark$                     &                       &                       & 83.67     & 79.18        \\
$\checkmark$                     & $\checkmark$                     &                       & 84.04     & 80.15        \\
$\checkmark$                     &                       & $\checkmark$                     & 84.33     & 80.23        \\
$\checkmark$                     & $\checkmark$                     & $\checkmark$                     & \textbf{85.12}     & \textbf{80.72}        \\
\hline
\end{tabular}
\end{table}

\begin{figure}[h]
\begin{minipage}[b]{.48\linewidth}
  \centering
    \includegraphics[width=4.0cm]{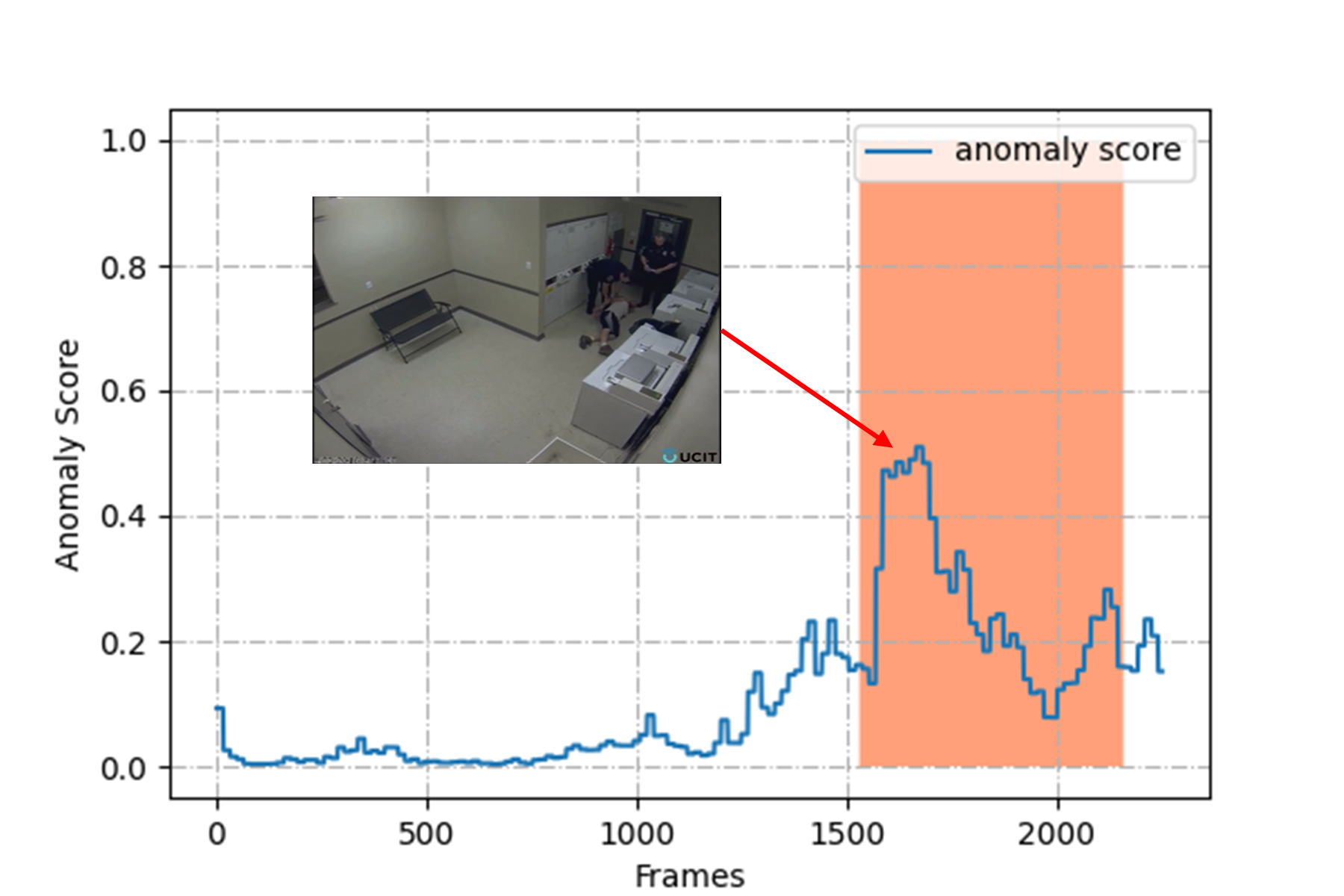}
  \centerline{(a) $\mathcal{L}_{MIL}$}\medskip
\end{minipage}
\hfill
\begin{minipage}[b]{0.48\linewidth}
  \centering
    \includegraphics[width=4.0cm]{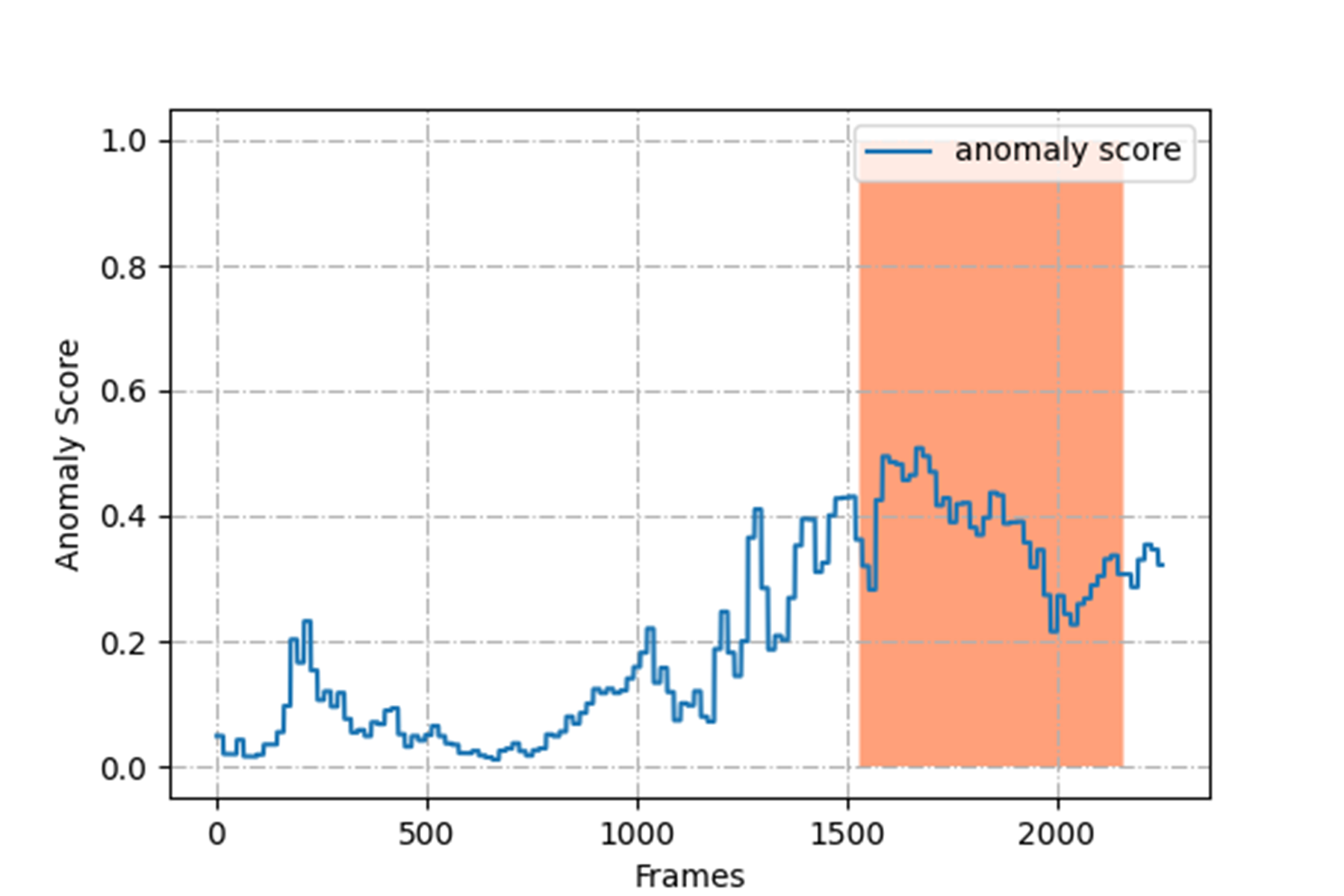}
  \centerline{(b) $\mathcal{L}_{MIL}$+$\mathcal{L}_{DR}$}\medskip
\end{minipage}
\begin{minipage}[b]{.48\linewidth}
  \centering
    \includegraphics[width=4.0cm]{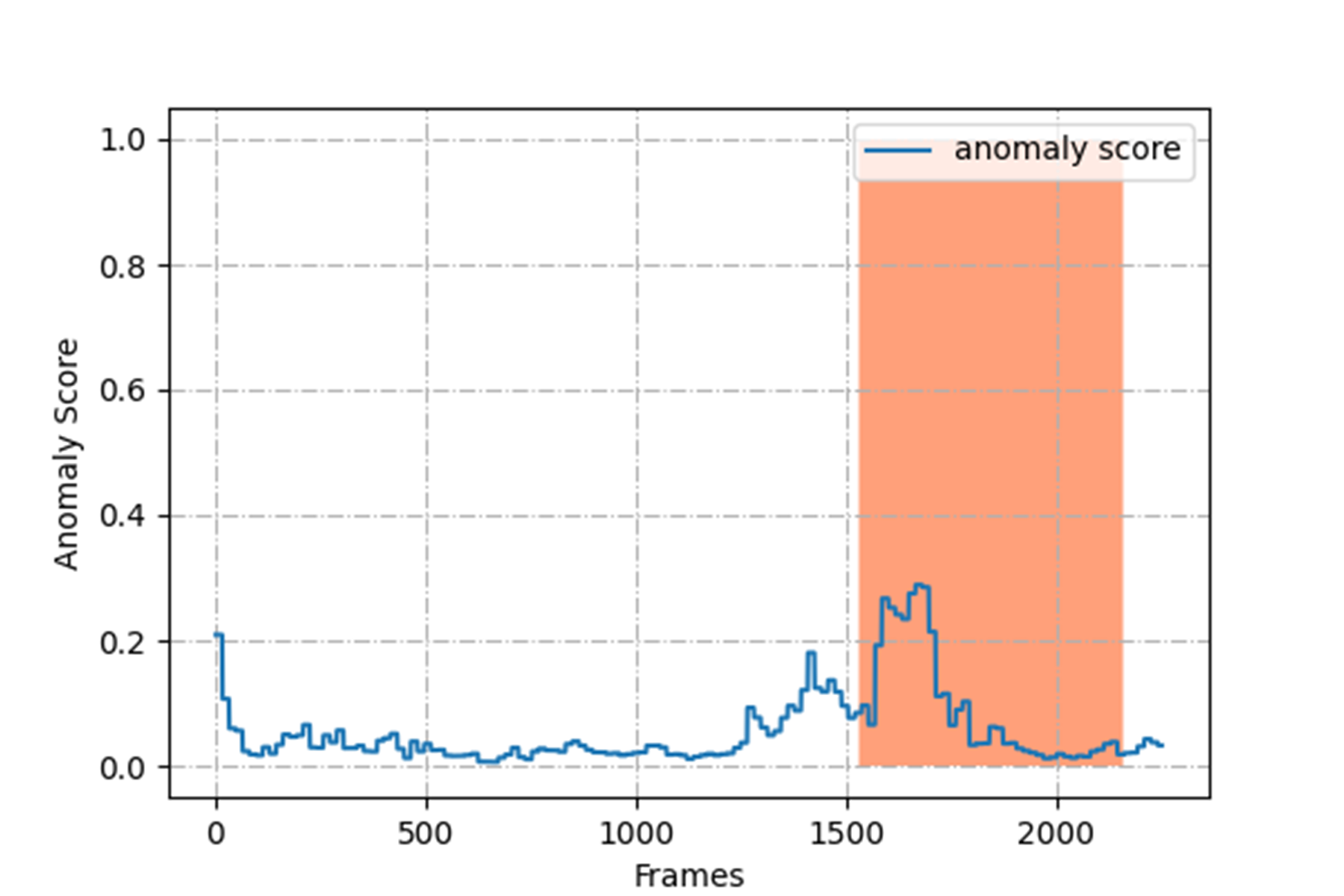}
  \centerline{(c) $\mathcal{L}_{MIL}$+$\mathcal{L}_{DA}$}\medskip
\end{minipage}
\hfill
\begin{minipage}[b]{0.48\linewidth}
  \centering
    \includegraphics[width=4.0cm]{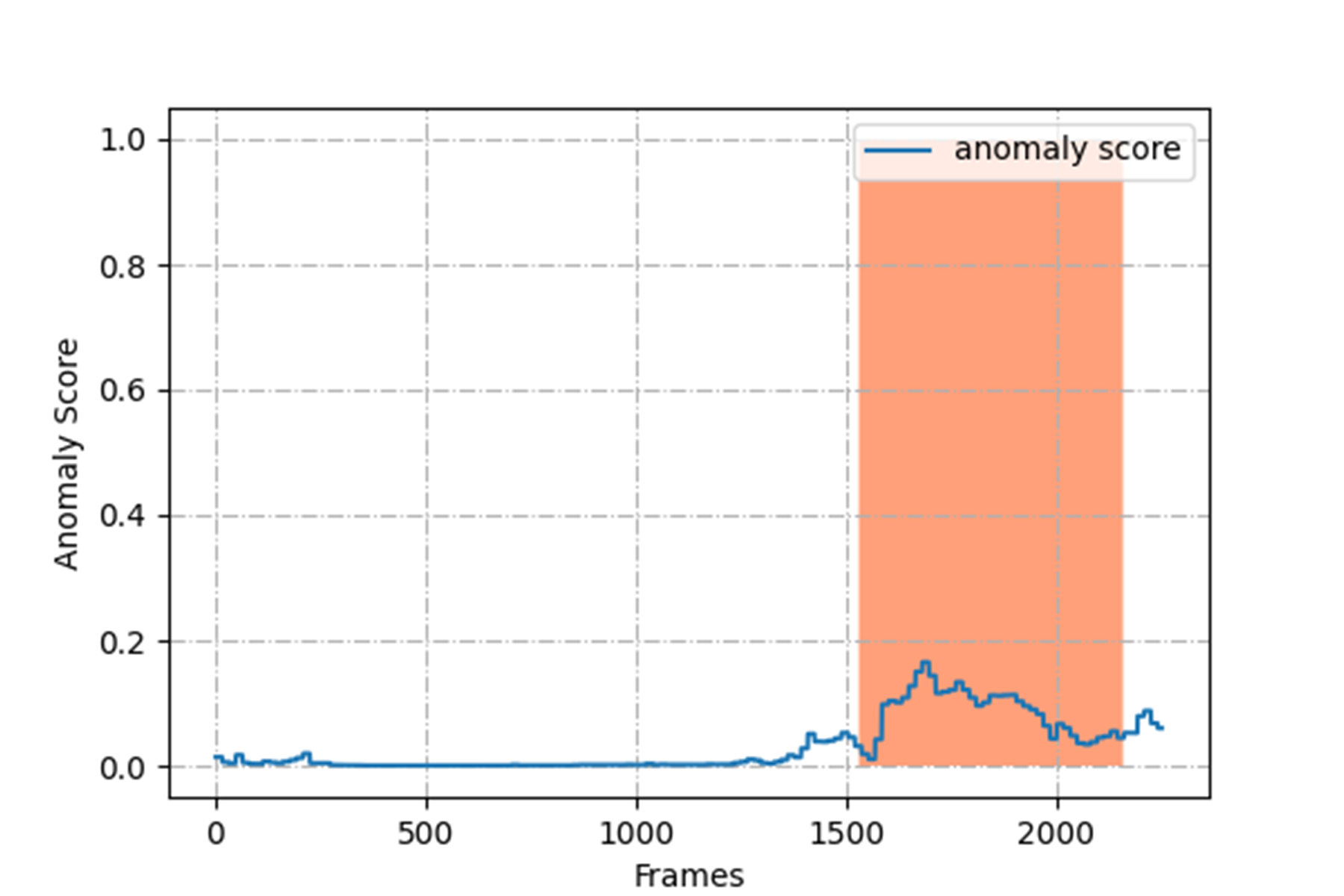}
  \centerline{(d) $\mathcal{L}_{MIL}$+$\mathcal{L}_{DR}$+$\mathcal{L}_{DA}$}\medskip
\end{minipage}
\caption{Visualization results of our DDL method on UCF-Crime. Orange windows indicate that the interval contains an abnormal event.}
\label{fig:res}
\end{figure}

\subsection{Qualitative Analysis}

Finally, we visualize the anomaly detection results before and after discriminative  dynamics learning. As shown in Fig.\,3(a), the anomaly score  predicted with MIL-based loss fluctuates widely before and after the ground truth boundary. With dynamics ranking loss, the anomaly score roughly increases over temporal dimension, as shown in Fig.\,3(b), which is due to the increase of score dynamics within the positive bag during the pairwise ranking process. Besides, it smoothes out the anomaly score within the ground truth interval. And after the temporal dynamics alignment, the peaks of the anomaly score in Fig.\,3(c) is suppressed and the overall prediction is in a lower range, thanks to the alignment implicitly improving the discriminability of the snippets within the bag. When both losses are added to the optimization, as Fig.\,3(d) shows, the predicted interval overlaps with the ground truth, and the anomaly score outside the boundary is significantly suppressed, which further improves the capacity to locate the anomaly intervals.

\section{Conclusion}
In this paper, we propose a Discriminative Dynamics Learning (DDL) method containing dynamics ranking loss and dynamics alignment loss for weakly supervised video anomaly detection. Dynamics ranking loss boosts the response magnitude of anomalous events by enlarging the anomaly score dynamics between positive and negative bags. Dynamics alignment loss explicitly aligns the two types of temporal dynamics within the video bag, thus smoothing out the dynamic variation of the inner bag snippets. In addition, we construct a prior-based Locality-aware Attention Network (LA-Net), which captures the long-range temporal dependencies across snippets while recalibrating the contextual correlations within the neighborhood. Experimental results on two large anomaly video datasets demonstrate the effectiveness of our approach. In the future, online detection and multimodal information will be further explored.

\section{ACKNOWLEDGEMENTS}
This work is supported by National Key R\&D Program of China (No.2021YFF0900701) and  National Natural Science Foundation of China (No.61801441, No. 61701277, No. 61771288).

\small
\bibliographystyle{IEEEbib}
\bibliography{icme2022template}

\end{document}